\pdfoutput=1
\documentclass[11pt]{article}

\usepackage[preprint]{acl}
\usepackage{times}
\usepackage{latexsym}
\usepackage[T1]{fontenc}
\usepackage[utf8]{inputenc}
\usepackage[ruled,vlined]{algorithm2e}
\usepackage{microtype}
\usepackage{inconsolata}
\usepackage{graphicx}
\usepackage{fancyvrb}
\usepackage{booktabs}
\usepackage{multirow}
\usepackage{amssymb}
\usepackage[most]{tcolorbox} 
\usepackage{enumitem}
\usepackage{fontawesome5}
\usepackage{mdframed}
\usepackage{verbatim}
\usepackage{stfloats}
\usepackage{listings}
\usepackage[table,xcdraw,dvipsnames]{xcolor}

\newcounter{promptctr}
\renewcommand{\thepromptctr}{Prompt~\Alph{promptctr}}

\definecolor{examplebg}{HTML}{e8f4ff}
\definecolor{examplebg2}{HTML}{F1F1EE}

\newtcolorbox{exampleblock}{
  enhanced,
  colback=examplebg2,
  colframe=examplebg2,   
  boxrule=0pt,
  arc=2mm,
  outer arc=2mm,
  left=8pt,
  right=8pt,
  top=6pt,
  bottom=6pt,
  boxsep=0pt,
  before skip=8pt,
  after skip=8pt,
  sharp corners=all,
  breakable
}

\tcbset{
  bw:domain/.style={
    colback=white,
    colframe=black!40,
    coltitle=black,
    fonttitle=\bfseries\sffamily,   
    colbacktitle=examplebg,
    boxrule=0.5pt,
    titlerule=0pt,
    arc=2pt,
    left=4pt,right=4pt,top=4pt,bottom=4pt,
    toptitle=2pt,bottomtitle=2pt,
  }
}

\newcommand{\mytcbinputwide}[5]{
  \begin{figure*}[t]
  \centering
  \refstepcounter{promptctr}
  \phantomsection
  \begin{tcolorbox}[title={\thepromptctr: #2},#4,width=\textwidth,enhanced]
    \lstinputlisting{#1}
    \label{#5}
  \end{tcolorbox}
  \vspace{-4pt}
  \end{figure*}
}

\newcommand{\promptref}[1]{%
  \hyperref[#1]{\ref*{#1}}%
}

\newcommand{\promptrefp}[1]{%
  \hyperref[#1]{\ref*{#1} (p.~\pageref*{#1})}%
}

\newcommand{\method}{\textsc{FD-NL2Sql}}

\title{\method{}: Feedback-Driven Clinical NL2SQL that Improves with Use}

\author{
\textsuperscript{1}Suparno Roy Chowdhury\thanks{These authors contributed equally.} \quad
\textsuperscript{1}Tejas Anvekar\footnotemark[1]  \quad
\textsuperscript{1}Manan Roy Choudhury\footnotemark[1] \\ 
\textbf{\textsuperscript{2}Muhammad Ali Khan} \quad 
\textbf{\textsuperscript{2}Kaneez Zahra Rubab Khakwani} \quad 
\textbf{\textsuperscript{2}Mohamad Bassam Sonbol} \\
\textbf{\textsuperscript{2}Irbaz Bin Riaz}\thanks{Corresponding authors.} \quad
\textbf{\textsuperscript{1}Vivek Gupta}\footnotemark[2] \\
  \raisebox{0.75ex}{\includegraphics[height=2ex]{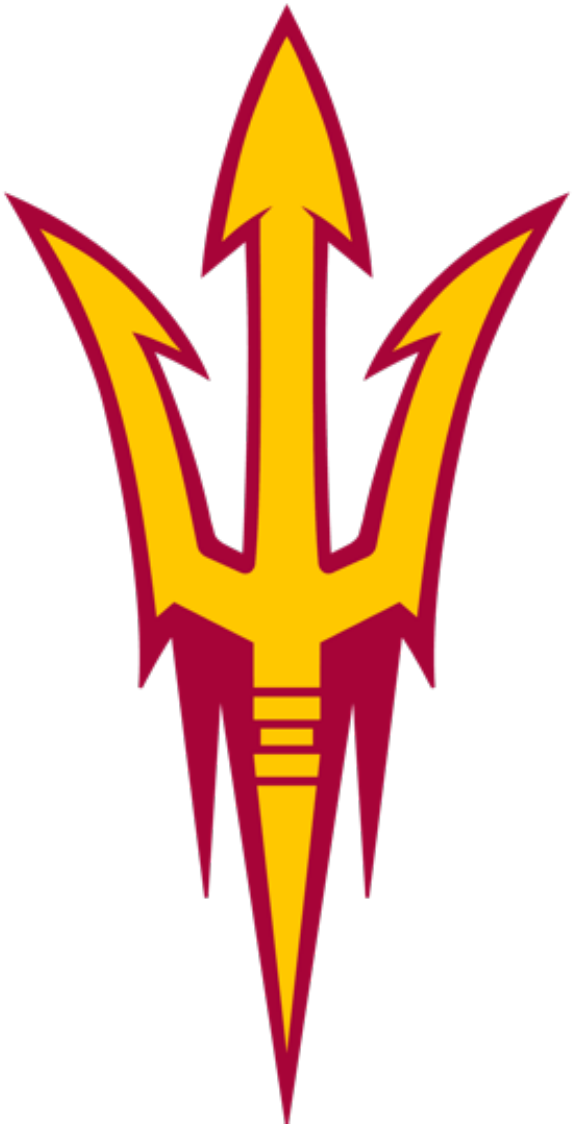}}\hspace{0.2ex}Arizona State University\textsuperscript{1} \quad   \raisebox{0.5ex}{\includegraphics[height=1.65ex]{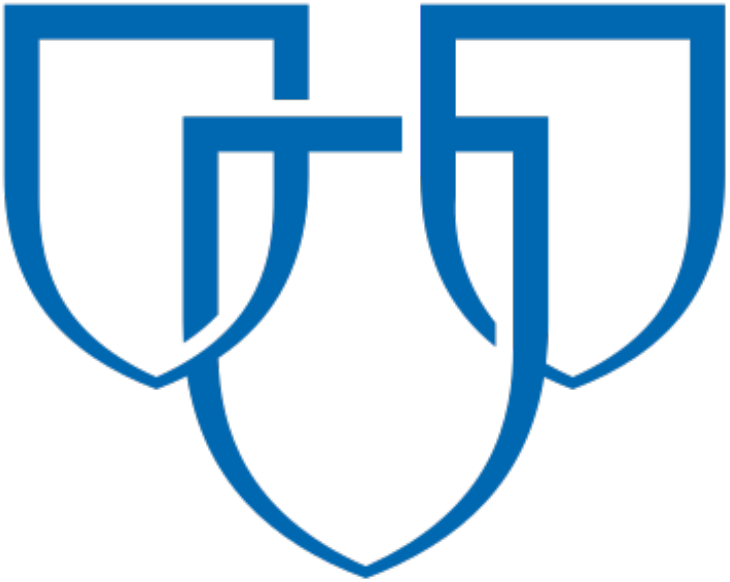}}\hspace{0.2ex}Mayo Clinic\textsuperscript{2} \\
\faGlobe~\href{https://tejasanvekar.github.io/FD-NL2SQL/}{Project Page} \quad
\faPlayCircle~\href{https://tejasanvekar.github.io/FD-NL2SQL/try}{Demo} \quad
\faVideo~\href{https://youtu.be/VMfOc440JKM}{Video} \quad
\faGithub~\href{https://github.com/TejasAnvekar/FD-NL2SQL}{Code} \\
\texttt{riaz.irbaz@mayo.edu, vgupt140@asu.edu}
}

\begin{document}
\maketitle

\begin{abstract}
Clinicians exploring oncology trial repositories often need ad-hoc, multi-constraint queries over biomarkers, endpoints, interventions, and time, yet writing SQL requires schema expertise. We demo \method{}, a feedback-driven clinical NL2SQL assistant for SQLite-based oncology databases. Given a natural-language question, a schema-aware LLM decomposes it into predicate-level sub-questions, retrieves semantically similar expert-verified NL2SQL exemplars via sentence embeddings, and synthesizes executable SQL conditioned on the decomposition, retrieved exemplars, and schema, with post-processing validity checks. To improve with use, \method{} incorporates two update signals: (i) clinician edits of generated SQL are approved and added to the exemplar bank; and (ii) lightweight logic-based SQL augmentation applies a single atomic mutation (e.g., operator or column change), retaining variants only if they return non-empty results. A second LLM generates the corresponding natural-language question and predicate decomposition for accepted variants, automatically expanding the exemplar bank without additional annotation. The demo interface exposes decomposition, retrieval, synthesis, and execution results to support interactive refinement and continuous improvement.
\end{abstract}

\section{Introduction}
\begin{figure}[t]
    \centering
    \includegraphics[width=0.95\linewidth]{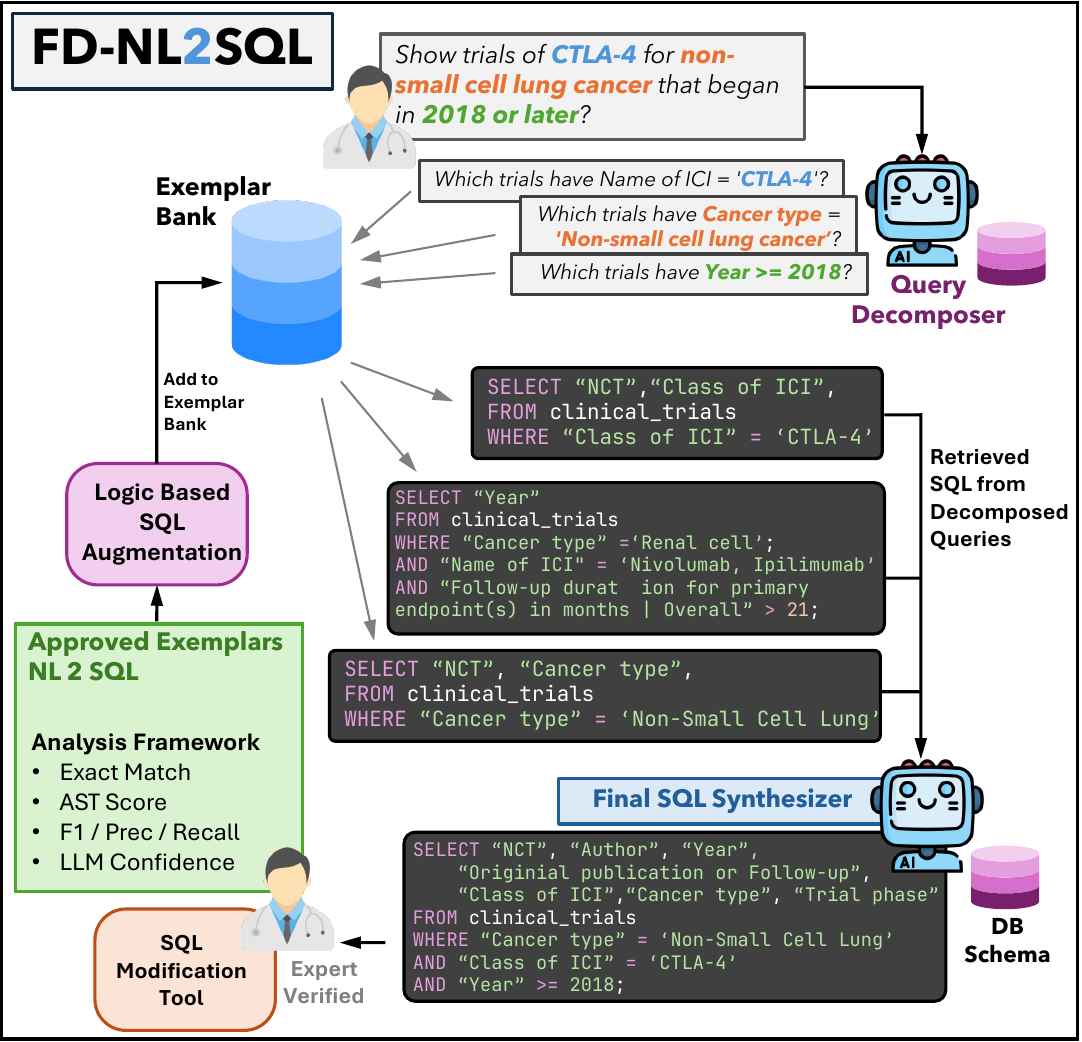}
    \vspace{-0.5em}
    \caption{A clinician question is decomposed into schema-aligned predicate sub-questions; for each predicate, semantically similar expert-approved exemplars are retrieved; This guides schema-grounded SQL synthesizer. Users can edit and approve the final SQL to update the exemplar bank. To expand coverage with minimal annotation, approved SQL is augmented by a single atomic mutation (e.g., operator or column substitution) and retained only if it returns non-empty results. A second LLM back-translates augmented SQL into a NL-question \& predicate sub-questions, new samples are added to the bank for continual improvement.}
    \label{fig:fdnl2sql}
    \vspace{-1.0em}
\end{figure}

Clinical trial databases are central to modern oncology research and drug development. Public registries such as ClinicalTrials.gov~\cite{zarin2011clinicaltrials} and institutional repositories contain rich structured data, including trial phase, biomarkers, eligibility criteria, endpoints, recruitment status, and sponsor information, supporting competitive intelligence, hypothesis generation, regulatory planning, and translational research. As oncology increasingly shifts toward biomarker-driven and precision trials, efficient access to this structured data is critical.

Yet these databases remain difficult to query. Access typically requires SQL expertise and detailed schema knowledge, Schemas are often complex, spanning multiple relational tables for eligibility, interventions, endpoints, \& disease ontologies. Clinicians and translational researchers, though domain experts, are rarely trained in database querying, leading to analyst-mediated workflows that slow iterative exploration. In the high-stakes setting of oncology drug development, this friction directly affects research velocity \& decision quality.

Existing tools only partially mitigate this gap. Registry interfaces rely on keyword-based search~\cite{zarin2011clinicaltrials}, which cannot reliably enforce structured multi-constraint filtering (e.g., biomarker + phase + endpoint + recruitment criteria). Business intelligence dashboards~\cite{chen2012business} provide predefined reports but remain rigid and cannot cover the combinatorial space of exploratory clinical queries.

Natural Language to SQL (NL2SQL) systems have advanced substantially in recent years, beginning with neural approaches such as Seq2SQL~\cite{zhong2017seq2sqlgeneratingstructuredqueries} \& large-scale benchmarks such as Spider~\cite{yu2018spider}. More recent work has introduced schema-aware reasoning (e.g., RAT-SQL~\cite{wang2020rat}) \& constrained decoding for syntactic validity (e.g., PICARD~\cite{scholak2021picard}). Large Language Models (LLMs) further demonstrate strong in-context semantic parsing capabilities~\cite{brown2020language}. However, these systems are largely designed for general-purpose benchmarks \& do not explicitly incorporate domain-aware constraint decomposition, exemplar retrieval grounded in clinical schemas, or interactive feedback loops tailored to high-stakes biomedical querying. In specialized domains such as oncology, naive generation without schema-aligned grounding \& domain-specific retrieval can lead to brittle or clinically implausible queries.

To address these limitations, we introduce a domain-aware NL2SQL system designed specifically for oncology clinical trial databases. Our approach integrates three core components. First, we perform LLM-guided self-evolving decomposition of a user’s question into atomic, schema-aligned sub-questions that each correspond to a filterable predicate. Second, we retrieve semantically similar seed exemplars using Sentence-BERT embeddings~\cite{reimers2019sentence}, enabling structured grounding in prior validated query patterns. Third, we perform retrieval-guided SQL synthesis with controlled decoding and post-processing to ensure structural validity and constraint satisfaction. This decomposition-retrieval-synthesis architecture improves robustness by aligning generation with both schema structure and domain-specific precedent.

Beyond static query translation, our system operates as a \emph{living clinical review assistant}. Each generated query can be previewed, refined, and corrected interactively. User feedback on retrieved exemplars and synthesized SQL is incorporated into the seed bank, improving retrieval neighborhoods and generation fidelity over time. This feedback-driven refinement is consistent with emerging paradigms of interactive and adaptive language model systems~\cite{brown2020language}, but is operationalized here in a structured, database-grounded clinical setting. As clinicians issue more domain-specific queries, the system progressively aligns with real-world oncology reasoning patterns.

From a clinician's perspective, this approach substantially reduces dependency on technical needs, accelerates hypothesis testing, \& enables real-time, multi-constraint exploration of trial criterias. By combining domain-aware decomposition, exemplar-guided synthesis, and iterative feedback, the system bridges the gap between oncology expertise and structured data access, transforming static registries into interactive analytical tools.

Finally, our main contributions are:
\begin{itemize}
    \item A schema-aware, predicate-level decomposition strategy that improves robustness of clinical NL2SQL in complex oncology databases.
    \item A retrieval-guided SQL synthesis pipeline that grounds LLM generation in expert-verified exemplars for reliable query construction.
    \item Feedback-driven, self-evolving clinical trial query assistant that improves continuously through clinician interaction.
\end{itemize}

\section{Related Work}
Recent advances in text-to-SQL have shifted from supervised semantic parsing toward large language model (LLM) prompting and in-context learning. DIN-SQL~\cite{10.5555/3666122.3667699} demonstrates that decomposed prompting improves SQL generation by breaking complex questions into intermediate reasoning steps. Similarly, execution-guided decoding~\cite{wang2018robusttexttosqlgenerationexecutionguided} improves robustness by validating generated queries against database constraints during generation. These approaches highlight the importance of structural grounding when generating executable SQL. Retrieval-based prompting has also emerged as an effective strategy for improving LLM reasoning. In-context example selection significantly affects downstream generation quality~\cite{liu-etal-2022-makes}, and retrieval-augmented generation (RAG)~\cite{NEURIPS2020_6b493230} shows that grounding outputs in external memory improves reliability. Our method extends this paradigm by retrieving semantically similar question-SQL exemplars using dense embeddings and conditioning synthesis on predicate-aligned decompositions, rather than relying solely on flat prompt demonstrations.

Within the biomedical domain, pretrained scientific language models such as SciBERT~\cite{beltagy-etal-2019-scibert} have demonstrated gains on domain-specific NLP tasks. However, prior work primarily focuses on unstructured text understanding rather than structured clinical database querying. Our system bridges biomedical language understanding with schema-aware SQL synthesis, enabling clinician-driven, multi-constraint exploration of oncology clinical trial databases.

\section{\method{}}
\subsection{System Architecture and Workflow}
\label{subsec:arch}
\definecolor{myYellow}{HTML}{f5c505}

\begin{figure*}[t]
    \centering
    \includegraphics[width=\linewidth]{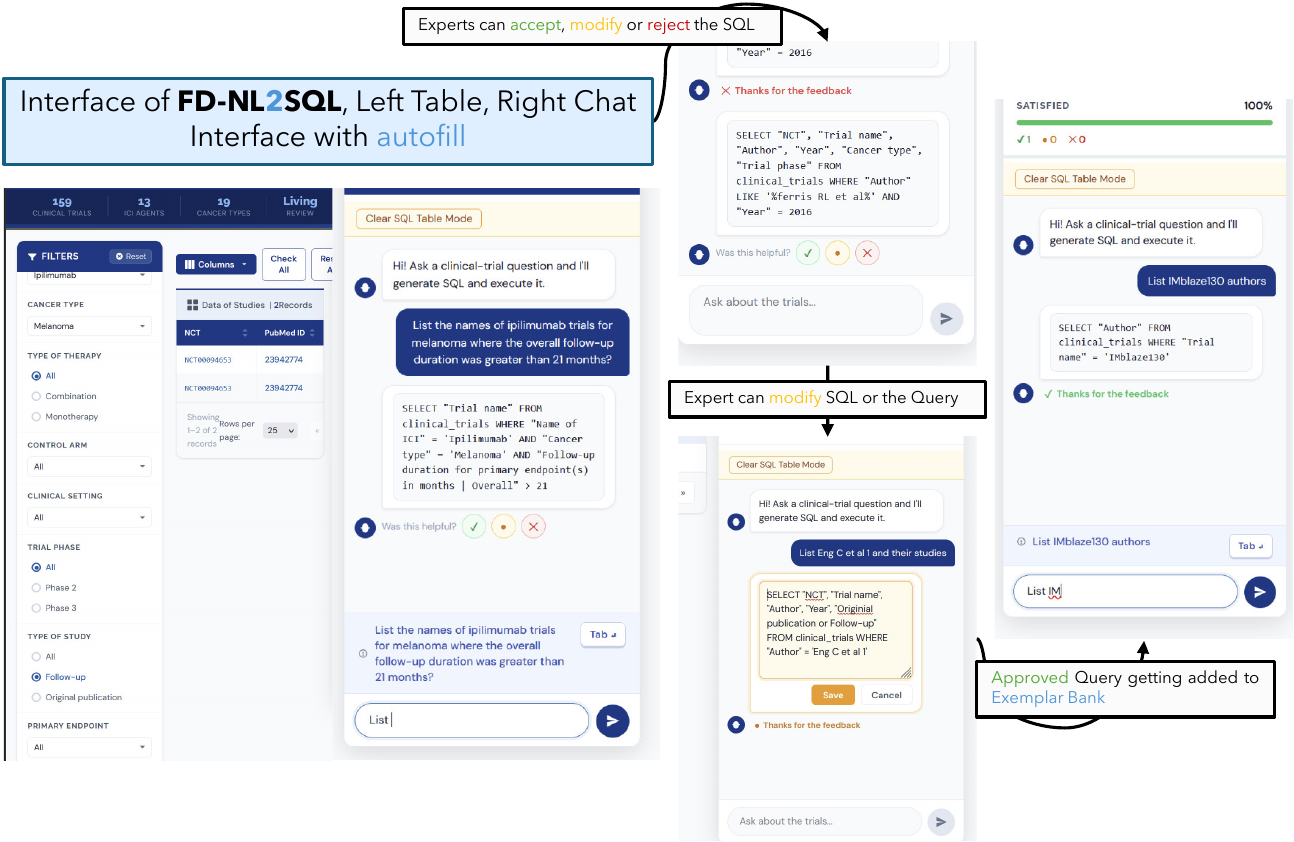}
    \vspace{-0.5em}
    \caption{\textsc{FD-NL2Sql} demo UI and feedback loop. Clinicians issue a natural-language query in the chat interface (right) and view executed results in the table view (left). The system shows the generated SQL, which an expert can \textcolor{ForestGreen}{\emph{accept}}, \textcolor{myYellow}{\emph{modify}}, or \textcolor{red}{\emph{reject}}; accepted / edited queries are saved back to the exemplar bank to improve future retrieval and synthesis (autofill supports rapid refinement).}
    \label{fig:fdnl2sql-UI}
    \vspace{-1.0em}
\end{figure*}

Illustrated in \autoref{fig:fdnl2sql}, \method{} is an interactive NL2SQL assistant for oncology clinical-trial databases. The system follows a modular pipeline combining LLM-based reasoning with retrieval over an evolving exemplar bank and lightweight programmatic checks. The demo UI exposes intermediate artifacts, retrieved exemplars, synthesized SQL, and results to support transparent refinement and feedback-driven
improvement.

\paragraph{Resources.}
We assume a SQLite database $\mathcal{D}$ with schema metadata and an exemplar bank $\mathcal{S}=\{(s_j,y_j)\}_{j=1}^{M}$ of expert-approved NL2SQL pairs. We pre-compute sentence embeddings $\mathbf{e}(s_j)$ for all $s_j$ and maintain an index for fast $topk$ retrieval.

\paragraph{1) Schema grounding.}
Before generation, we introspect $\mathcal{D}$ to build a schema dictionary (tables, columns, types, and join keys). This schema context is injected into prompts and used for post-generation validation (e.g., column existence and join feasibility).

\paragraph{2) Decomposition-retrieval.}
Given a user question $x$, an LLM produces a schema-aligned, \texttt{WHERE}-oriented decomposition $\mathcal{X}(x)=\{x_1,\ldots,x_n\}$ where each $x_i$ targets one atomic predicate (column, operator, value). For each $x_i$, we retrieve the top-$k_r$ nearest exemplars from $\mathcal{S}$ using cosine similarity in embedding space, storing $(s_j,y_j,\mathrm{score}(x_i,s_j))$. To reduce prompt noise, we additionally extract a compact \texttt{WHERE}-pattern hint from each retrieved SQL when available.

\paragraph{3) Retrieval-guided SQL synthesis \& execution.}
A second LLM synthesizes the final SQL $\hat{y}$ conditioned on (i) $x$, (ii) $\mathcal{X}(x)$, and (iii) the retrieved exemplar bundles $\{(x_i,\mathcal{N}_i)\}_{i=1}^{n}$. The model is instructed to use retrieved SQL as structural templates while satisfying all constraints in $x$ and avoiding irrelevant literal copying. We parse the output into a single executable \texttt{SELECT} / \texttt{WITH} statement, apply lightweight guards (read-only policy, schema checks, timeout), and execute $\hat{y}$ against SQLite to render results in the UI.

\subsection{Feedback and Exemplar Bank Expansion}
\label{subsec:feedback}

\paragraph{4) Expert approval.}
The \autoref{fig:fdnl2sql-UI} interface allows users to edit generated SQL. If an expert approves the corrected
query $y^{\star}$ for question $x$, the pair $(x,y^{\star})$ is appended to the exemplar
bank $\mathcal{S}$ and embedded for future retrieval.

\paragraph{5) SQL augmentation for bank growth.}
To expand exemplar coverage with minimal manual annotation, \method{} augments approved queries in two sub-steps:

\begin{itemize}
    \item[a)] \textbf{Logic-based SQL mutation.}
Starting from an approved query $y^{\star}$, we apply exactly one atomic transformation, such as (i) operator change (\texttt{=} $\rightarrow$ \texttt{>=}, \texttt{LIKE}, etc.), (ii) column substitution within a compatible type group, or (iii) controlled value edits (e.g., year thresholds). The mutated query $\tilde{y}$ is retained only if it executes
successfully and returns a non-empty result on $\mathcal{D}$; otherwise it is discarded.

 \item[b)] \textbf{NL Back-Translation}
For each retained $\tilde{y}$, a separate LLM generates (i) a natural-language question $\tilde{x}$ consistent with $\tilde{y}$ and (ii) predicate-level sub-questions $\mathcal{X}(\tilde{x})$ aligned with the mutated constraints. The resulting pair $(\tilde{x},\tilde{y})$ (and optional decomposition) is added back to $\mathcal{S}$, improving retrieval and synthesis coverage over time.

\end{itemize}

\section{Experimental Setup}
\label{sec:exp}

\subsection{Dataset Construction}
\label{subsec:data}

\paragraph{Seed set.}
We start with 500 seed questions authored by a Mayo Clinic oncology scientist to reflect realistic evidence-review queries over the IOTOX\footnote{https://iotox.living-evidence.com/} table (e.g., cancer type, class of ICI, trial phase, endpoints / follow-up, temporal filters). Each question is paired with a gold SQLite query, verified by execution, and stored as JSON (question, SQL, optional metadata). The seed set serves as (i) the initial exemplar bank for retrieval and (ii) the pool for few-shot demonstrations.

\paragraph{Programmatic benchmark expansion.}
To evaluate generalization beyond the seed distribution, we create  benchmark by applying a \emph{single} atomic transformation to each seed pair $(x,y): (\tilde{x},\tilde{y}) = T(x,y)$. where $T$ edits either the projection (\texttt{SELECT}) or constraints (\texttt{WHERE}) while preserving the overall query intent. We generate the following variant types (counts in parentheses): \textbf{1) Projection edits:} keep only 2 selected columns (275); drop one selected column (238). \textbf{2) Predicate dropping:} drop one \texttt{WHERE} condition (179); keep only one \texttt{WHERE} condition (115); drop 2 \texttt{WHERE} conditions (99); remove \texttt{WHERE} entirely (20). \textbf{3) Value edits:} swap a text equality value (40); relax a numeric threshold (34).

\paragraph{Validity filtering and de-duplication.}
Each generated SQL $\tilde{y}$ is executed against SQLite. We retain a variant only if it (i) is read-only (\texttt{SELECT}/\texttt{WITH}), (ii) executes without error, and (iii) returns a non-empty result. We then normalize and de-duplicate SQL to remove repeats. Finally, we use an open-source LLM \texttt{Gemma3-27B-it}~\cite{gemma_2025} to generate natural-language query using \promptref{prompt:SQL2NL} for each augmented SQL. This constituted to 1500 NL2SQL samples \& authors of this paper verified its correctness.
    
\subsection{Baselines and Benchmark Settings}
\label{subsec:baselines}

We compare \method{} against four representative NL2SQL baselines:
1) zero-shot, 2) few-shot, 3) Chain-of-Thoughts, using both open-source and closed source LLMs (\texttt{Gemma3-27B-it}~\cite{gemma_2025}, \texttt{Qwen3-30B-A3B-Instruct-2507}~\cite{qwen3}, and \texttt{gpt-5-nano/mini}~\cite{gpt-5}) and 4) \texttt{SQL-R1-14B}~\cite{sqlr1} finetuned reasoning based model trained with reinforcement-learning objectives. We use deterministic decoding where supported (e.g., temperature 0.0). All prompts of baselines (\promptref{prompt:ZS}, \promptref{prompt:FS}, \promptref{prompt:CoT}) and our method (\promptref{prompt:decompose}, \promptref{prompt:synth}) are given in \autoref{sec:appendix_prompts}.

\subsection{Evaluation Metrics}
\label{subsec:metrics}

We report execution-based correctness on the same SQLite database, plus lightweight structural / text scores. Specifically, we compute: (i) \textbf{eEM} (exact match of predicted vs.\ gold result multisets after column alignment and value canonicalization), (ii) \textbf{eF1} (soft row-overlap F1 after one-to-one row matching with numeric tolerance and token overlap for strings), (iii) \textbf{\textsc{chrF}} (computed on the matched (row-aligned) predicted vs.\ gold result strings) (iv) \textbf{AST}\footnote{\href{https://continuous-eval.docs.relari.ai/metrics/code/deterministic/sql_ast_similarity/}{sqlglot}} (clause-weighted token-F1 over \texttt{SELECT} / \texttt{FROM} / \texttt{WHERE} with weights $(w_s,w_w,w_f)=(0.5,0.4,0.1)$ when parsing succeeds), (v) LLM \textbf{Conf} from \texttt{structured\_logprobs}\footnote{\href{https://arena-ai.github.io/structured-logprobs/}{structured-logprobs}} over the generated SQL tokens, and  We additionally log diagnostic flags for non-read-only SQL, multiple statements, and \texttt{LIMIT} without \texttt{ORDER BY}.

\section{Results and Discussions}
In this section, we evaluate the performance of FD-NL2SQL and analyze its behavior across realistic oncology clinical trial queries. We report quantitative results on execution-based metrics and examine how decomposition, retrieval grounding, and feedback-driven augmentation contribute to robustness and continuous improvement. We further discuss practical implications and observed limitations in clinician-facing usage.

\begin{table}[!htbp]
\caption{\textbf{Quantitative results} across prompting strategies, finetuned model and \textsc{FD-NL2Sql}(Ours). We report \textsc{chrF}, eEM, eF1, AST, and Conf; \textit{HM} denotes the harmonic mean of \{Conf, eF1, eEM\}. \textcolor{ForestGreen}{Green} marks the best score \emph{within a given strategy for each model}, \textcolor{red}{red} marks the worst score within that strategy for each model, and \underline{underlines} denote the best score overall across all models and strategies for a metric. Note \texttt{gpt-5n} depicts \texttt{nano}, and \texttt{gpt-5m} for \texttt{mini} variant}
\label{tab:results}
\centering
\setlength{\tabcolsep}{3pt}
\resizebox{\columnwidth}{!}{%
\begin{tabular}{llcccccc}
\toprule
\textbf{Strategy} &
  \textbf{Models} &
  \multicolumn{1}{c}{\textbf{\textsc{chrF}}} &
  \multicolumn{1}{c}{\textbf{eEM}} &
  \multicolumn{1}{c}{\textbf{eF1}} &
  \multicolumn{1}{c}{\textbf{AST}} &
  \multicolumn{1}{c}{\textbf{Conf}} &
  \multicolumn{1}{c}{\textbf{\textit{HM}}} \\ \midrule
\multirow{5}{*}{Zero-Shot}          & \texttt{Qwen3}      & 81.82 & 26.90 & 46.51 & \textcolor{red}{72.54} & 39.01 & 51.58 \\
                                    & \texttt{Gemma3}     & 84.07 & 29.48 & 39.95 & 59.41 & 65.69 & 49.95 \\
                                    & \texttt{SQL-R1}     & 63.78 & 03.32 & \textcolor{red}{01.75} & \textcolor{red}{41.78} & 10.39 & 00.42 \\
                                    & \texttt{gpt-5n} & 89.13 & \textcolor{red}{11.60} & 18.87 & \textcolor{red}{52.58} & -- & -- \\
                                    & \texttt{gpt-5m} & \textcolor{ForestGreen}{\textbf{\underline{94.64}}} & 38.33 & \textcolor{red}{47.36} & \textcolor{red}{51.21} & -- & -- \\ \midrule
\multirow{4}{*}{Few-Shot}           & \texttt{Qwen3}      & 83.02 & 25.60 & 52.95 & 80.43 & 34.70 & 53.15 \\
                                    & \texttt{Gemma3}     & 91.32 & 30.85 & \textcolor{ForestGreen}{\textbf{45.72}} & 85.47 & 53.16 & 56.80 \\
                                    & \texttt{gpt-5n} & 89.58 & 15.80 & 31.97 & 75.53 & -- & -- \\
                                    & \texttt{gpt-5m} & 94.52 & 31.27 & 46.85 & 83.03 & -- & -- \\ \midrule
\multirow{5}{*}{CoT}                & \texttt{Qwen3}      & \textcolor{red}{77.96} & \textcolor{red}{21.65} & \textcolor{red}{29.73} & 85.21 & \textcolor{red}{00.00} & \textcolor{red}{00.00} \\
                                    & \texttt{Gemma3}     & \textcolor{red}{77.89} & \textcolor{red}{13.47} & \textcolor{red}{34.88} & \textcolor{red}{00.00} & \textcolor{red}{00.25} & \textcolor{red}{00.18} \\
                                    & \texttt{SQL-R1}     & \textcolor{red}{50.24} & \textcolor{red}{02.85} & 03.35 & 49.15 & \textcolor{red}{01.69} & \textcolor{red}{00.15} \\
                                    & \texttt{gpt-5n} & \textcolor{red}{86.93} & 13.67 & \textcolor{red}{33.47} & 77.56 & -- & -- \\
                                    & \texttt{gpt-5m} & \textcolor{red}{86.28} & \textcolor{red}{10.33} & \textcolor{red}{19.87} & 82.02 & -- & -- \\ \midrule
\multirow{4}{*}{Ours}      
                                    & \texttt{Qwen3}      & \textcolor{ForestGreen}{\textbf{92.02}} & \textcolor{ForestGreen}{\textbf{\underline{40.93}}} & \textcolor{ForestGreen}{\textbf{\underline{55.57}}} & \textcolor{ForestGreen}{\textbf{86.36}} & \textcolor{ForestGreen}{\textbf{66.30}} & \textcolor{ForestGreen}{\textbf{\underline{66.06}}} \\
                                    & \texttt{Gemma3}     & \textcolor{ForestGreen}{\textbf{92.62}} & \textcolor{ForestGreen}{\textbf{32.60}} & 44.48 & \textcolor{ForestGreen}{\textbf{86.21}} & \textcolor{ForestGreen}{\textbf{\underline{70.48}}} & \textcolor{ForestGreen}{\textbf{57.88}} \\
                                    & \texttt{gpt-5n} & \textcolor{ForestGreen}{\textbf{90.89}} & \textcolor{ForestGreen}{\textbf{32.53}} & \textcolor{ForestGreen}{\textbf{50.25}} & \textcolor{ForestGreen}{\textbf{86.32}} & -- & -- \\
                                    & \texttt{gpt-5m} & 92.68 & \textcolor{ForestGreen}{\textbf{39.20}} & \textcolor{ForestGreen}{\textbf{55.85}} & \textcolor{ForestGreen}{\textbf{\underline{87.14}}} & -- & -- \\ \bottomrule 
\end{tabular}%
}
\end{table}

\paragraph{RQ1: Why does \textsc{FD-NL2Sql} perform better?}
\autoref{tab:results} shows that \textsc{FD-NL2Sql} consistently improves execution-based correctness over standard prompting (zero-shot, few-shot, CoT) across all backbone models, with the largest gains for backbones such as \texttt{Qwen3} and \texttt{gpt-5-nano}. The gains are reflected across complementary metrics: \textbf{eEM} increases when the predicted query returns the \emph{exact} gold result set; \textbf{eF1} increases when partial correctness improves (e.g., more constraints are satisfied even if not all); and higher \textbf{AST} indicates the generated SQL follows the gold structure more closely, reducing schema / logic drift. This translates into stronger \textbf{HM} scores, indicating that \textsc{FD-NL2Sql} improves not just one metric but the overall balance of correctness, structure, and confidence.

Prompting-only baselines are brittle for multi-constraint clinical questions: zero-shot frequently drops predicates or selects incompatible columns, few-shot remains sensitive to which demonstrations are shown, and CoT often degrades execution by introducing extra noise and incorrect intermediate assumptions that leak into the final SQL. The reasoning-finetuned baseline \texttt{SQL-R1} performs particularly poorly in our setting, which we attribute to domain/schema mismatch: oncology trial tables use domain-specific fields and naming conventions, so plausible-but-ungrounded SQL (guessed columns, joins, or values) can parse yet execute to the wrong results. \textsc{FD-NL2Sql} addresses these failure modes by enforcing \emph{schema-aligned predicate decomposition} (reducing omission) and \emph{in-domain exemplar grounding} (retrieved expert-approved templates), which together stabilize generation and improve both execution and structure.

\paragraph{RQ2: Why such tool is necessary in medical domain?}
Clinical evidence review workflows routinely involve exploratory, ad-hoc querying over structured trial attributes (e.g., cancer type, checkpoint inhibitors, endpoint, follow-up windows, inclusion criteria). However, operationalizing these questions against a database requires SQL proficiency \emph{and} detailed schema familiarity; an unrealistic expectation for clinicians \& researchers operating under time constraints. Baseline results shows that \emph{``just prompt an LLM''} is not a dependable substitute: even strong general models can produce SQL that is partially correct (moderate \textsc{chrF} / eF1) but fails exact execution (lower eEM), which is problematic when decisions depend on precise filtering and reproducible counts. Moreover, clinical trial databases evolve: columns change, new values appear, and conventions shift, making static prompt templates and one-off engineering brittle. This motivates a system that reduces the expertise barrier while remaining robust to schema complexity and domain-specific terminology.

\paragraph{RQ3: What is the practical utility of \textsc{FD-NL2Sql} beyond leaderboard?}
The primary objective is \emph{executable SQL}, which is inherently reproducible, and auditable  properties that matter in medical workflows more than a natural-language answer alone. The system’s trace (decomposition $\rightarrow$ retrieved exemplars $\rightarrow$ synthesized SQL $\rightarrow$ results) supports interactive refinement: users can see which constraints were extracted, which prior exemplars influenced the query, and what exactly will be executed. This directly addresses common clinician needs such as \emph{``tighten this criterion,''} \emph{``drop this filter,''} or \emph{``change the endpoint window,''} without requiring users to learn schema details from scratch. The results table also suggests an important practical pattern: \textsc{FD-NL2Sql} narrows performance gaps between backbones, enabling smaller/cheaper models to approach the reliability of larger models when paired with decomposition and exemplar guidance. In deployed settings, this can translate into lower latency and cost while preserving usability.

\paragraph{RQ4: How does \textsc{FD-NL2Sql} improve reliability and trust?}
Reliability in medical-domain NL2SQL is not only about average accuracy, but about avoiding silent failures and enabling verification. We therefore emphasize execution-based metrics (eEM / eF1) and structural checks (ASTSim), and we log diagnostic flags for unsafe/non-read-only SQL, multiple statements, and \texttt{LIMIT} without \texttt{ORDER BY}. These safeguards reduce the chance that a model output is executed in an unsafe or misleading way. In addition, we report \textbf{LLM confidence}, which provides a likelihood signal that can be used to triage queries for review (e.g., low-confidence generations may warrant user confirmation or stronger retrieval evidence). Finally, the feedback loop makes the system more dependable over time: expert-approved SQL is added to the exemplar bank, and single-step SQL mutations are retained only if they execute and return non-empty results before being back-translated into new NL2SQL exemplars. This combination of human approval, database-grounded filtering, and transparent intermediate artifacts supports a ``living'' assistant that can adapt to evolving clinical trials while remaining inspectable and reproducible.

\section{Conclusion}
\label{sec:conclusion}

We presented \textsc{FD-NL2Sql}, a feedback-driven clinical NL2SQL demo system for oncology trial databases that couples schema-aligned predicate decomposition with exemplar-guided SQL synthesis. Across multiple backbone models, \textsc{FD-NL2Sql} improves execution-based correctness and structural fidelity over standard prompting, demonstrating that retrieval grounding and constraint-level decomposition are more reliable than prompting alone for multi-constraint clinical queries. Beyond accuracy, the system is designed for real evidence-review workflows: it produces executable, auditable SQL; exposes an interpretable trace (decomposition, retrieved exemplars, final query, results); and supports expert-in-the-loop refinement. Finally, \textsc{FD-NL2Sql} continuously improves with use by incorporating approved queries and by safely expanding its exemplar bank via single-step SQL mutations validated  and LLM back-translation. Together, this enables a practical ``living'' assistant that lowers the barrier to structured trial exploration while maintaining transparency and reproducibility required in the medical domain.

\section*{Acknowledgment}
This research was supported by the Mayo Clinic and Arizona State University Alliance for Health Care Collaborative Research Seed Grant Program (Award ID: AWD00041508; Sponsor Award ID: ARI-358187) for the project: `Artificial intelligence-assisted digital, living, interactive clinical practice guidelines for cancer providers and patients.’

\section*{Limitations}
Our benchmark is derived from 500 expert-authored seed questions with controlled single-edit variants (projection/WHERE/value edits), which enables targeted stress testing but may not capture the full diversity of real clinician queries (e.g., complex joins, nested SQL, richer aggregations, or broader linguistic variation) and can introduce distributional biases toward simplified edits. Our exemplar expansion retains only augmented SQL that executes and returns non-empty results, which may over-represent frequent conditions and under-cover rare cancers or edge-case endpoints; moreover, the augmentation operators are intentionally lightweight (single atomic mutations), limiting automatic generation of more complex query patterns. Finally, confidence via \texttt{structured\_logprobs} is not uniformly available across model APIs, and we evaluate offline SQL correctness without prospective user studies or formal deployment privacy / security assessments.

\section*{Ethics Statement}

FD-NL2SQL is designed to support clinician-facing exploration of oncology clinical trial repositories for research and evidence review, not for patient-specific diagnosis or treatment decisions. The system operates on publicly available Mayo Clinic clinical trial data and a SQLite representation derived from it. The dataset contains no patient-level records, protected health information (PHI), or personally identifiable information (PII). Seed questions used to initialize the exemplar bank were authored by Mayo Clinic collaborators to reflect realistic clinical information needs and contain no sensitive data.

Because large language models may generate incorrect or incomplete SQL, the system prioritizes transparency and safety: it outputs executable, auditable SQL rather than free-form text, exposes intermediate reasoning steps (decomposition, retrieved exemplars, and synthesized query), and enforces lightweight execution safeguards including read-only query constraints, schema validation, and timeout checks. The system incorporates a feedback-driven update mechanism in which only expert-approved SQL edits are added to the exemplar bank, reducing the risk of quality degradation. Automatic SQL augmentation retains variants only if they execute successfully and return non-empty results, though this may bias coverage toward more frequent query patterns.

While the underlying data is public and non-identifiable, any future institutional deployment should include access controls, audit logging, and safeguards discouraging the entry of sensitive information into free-text interfaces. The system is intended to augment, not replace, expert clinical judgment. Portions of the manuscript text were edited and polished with the assistance of AI-based language tools, with all technical content and claims reviewed and validated by the authors.

\bibliography{anthology}

@article{zarin2011clinicaltrials,
    author = {Zarin, Deborah A and Tse, Tony and Williams, Rebecca J and Califf, Robert M and Ide, Nicholas C},
    address = {WALTHAM},
    copyright = {Copyright © 2011 Massachusetts Medical Society. All rights reserved.},
    issn = {0028-4793},
    journal = {The New England journal of medicine},
    language = {eng},
    number = {9},
    pages = {852-860},
    publisher = {Massachusetts Medical Society},
    title = {The ClinicalTrials.gov Results Database — Update and Key Issues},
    volume = {364},
    year = {2011},
}

@article{chen2012business,
    author = {Chen, Hsinchun and Chiang, Roger H. L. and Storey, Veda C.},
    address = {MINNEAPOLIS},
    copyright = {Copyright © 2012 Management Information Systems Research Center (MISRC) of the University of Minnesota},
    issn = {0276-7783},
    journal = {MIS quarterly},
    language = {eng},
    number = {4},
    pages = {1165-1188},
    publisher = {Management Information Systems Research Center, University of Minnesota},
    title = {Business Intelligence and Analytics: From Big Data to Big Impact},
    volume = {36},
    year = {2012},
}

@misc{zhong2017seq2sqlgeneratingstructuredqueries,
      title={Seq2SQL: Generating Structured Queries from Natural Language using Reinforcement Learning}, 
      author={Victor Zhong and Caiming Xiong and Richard Socher},
      year={2017},
      eprint={1709.00103},
      archivePrefix={arXiv},
      primaryClass={cs.CL},
      url={https://arxiv.org/abs/1709.00103}, 
}

@inproceedings{yu2018spider,
    title = "{S}pider: A Large-Scale Human-Labeled Dataset for Complex and Cross-Domain Semantic Parsing and Text-to-{SQL} Task",
    author = "Yu, Tao  and
      Zhang, Rui  and
      Yang, Kai  and
      Yasunaga, Michihiro  and
      Wang, Dongxu  and
      Li, Zifan  and
      Ma, James  and
      Li, Irene  and
      Yao, Qingning  and
      Roman, Shanelle  and
      Zhang, Zilin  and
      Radev, Dragomir",
    editor = "Riloff, Ellen  and
      Chiang, David  and
      Hockenmaier, Julia  and
      Tsujii, Jun{'}ichi",
    booktitle = "Proceedings of the 2018 Conference on Empirical Methods in Natural Language Processing",
    month = oct # "-" # nov,
    year = "2018",
    address = "Brussels, Belgium",
    publisher = "Association for Computational Linguistics",
    url = "https://aclanthology.org/D18-1425/",
    doi = "10.18653/v1/D18-1425",
    pages = "3911--3921"
}

@inproceedings{wang2020rat,
    title = "{RAT-SQL}: Relation-Aware Schema Encoding and Linking for Text-to-{SQL} Parsers",
    author = "Wang, Bailin  and
      Shin, Richard  and
      Liu, Xiaodong  and
      Polozov, Oleksandr  and
      Richardson, Matthew",
    editor = "Jurafsky, Dan  and
      Chai, Joyce  and
      Schluter, Natalie  and
      Tetreault, Joel",
    booktitle = "Proceedings of the 58th Annual Meeting of the Association for Computational Linguistics",
    month = jul,
    year = "2020",
    address = "Online",
    publisher = "Association for Computational Linguistics",
    url = "https://aclanthology.org/2020.acl-main.677/",
    doi = "10.18653/v1/2020.acl-main.677",
    pages = "7567--7578"
}

@inproceedings{scholak2021picard,
    title = "{PICARD}: Parsing Incrementally for Constrained Auto-Regressive Decoding from Language Models",
    author = "Scholak, Torsten  and
      Schucher, Nathan  and
      Bahdanau, Dzmitry",
    editor = "Moens, Marie-Francine  and
      Huang, Xuanjing  and
      Specia, Lucia  and
      Yih, Scott Wen-tau",
    booktitle = "Proceedings of the 2021 Conference on Empirical Methods in Natural Language Processing",
    month = nov,
    year = "2021",
    address = "Online and Punta Cana, Dominican Republic",
    publisher = "Association for Computational Linguistics",
    url = "https://aclanthology.org/2021.emnlp-main.779/",
    doi = "10.18653/v1/2021.emnlp-main.779",
    pages = "9895--9901"
}

@inproceedings{reimers2019sentence,
    title = "Sentence-{BERT}: Sentence Embeddings using {S}iamese {BERT}-Networks",
    author = "Reimers, Nils  and
      Gurevych, Iryna",
    editor = "Inui, Kentaro  and
      Jiang, Jing  and
      Ng, Vincent  and
      Wan, Xiaojun",
    booktitle = "Proceedings of the 2019 Conference on Empirical Methods in Natural Language Processing and the 9th International Joint Conference on Natural Language Processing (EMNLP-IJCNLP)",
    month = nov,
    year = "2019",
    address = "Hong Kong, China",
    publisher = "Association for Computational Linguistics",
    url = "https://aclanthology.org/D19-1410/",
    doi = "10.18653/v1/D19-1410",
    pages = "3982--3992"
}

@inproceedings{brown2020language,
 author = {Brown, Tom and Mann, Benjamin and Ryder, Nick and Subbiah, Melanie and Kaplan, Jared D and Dhariwal, Prafulla and Neelakantan, Arvind and Shyam, Pranav and Sastry, Girish and Askell, Amanda and Agarwal, Sandhini and Herbert-Voss, Ariel and Krueger, Gretchen and Henighan, Tom and Child, Rewon and Ramesh, Aditya and Ziegler, Daniel and Wu, Jeffrey and Winter, Clemens and Hesse, Chris and Chen, Mark and Sigler, Eric and Litwin, Mateusz and Gray, Scott and Chess, Benjamin and Clark, Jack and Berner, Christopher and McCandlish, Sam and Radford, Alec and Sutskever, Ilya and Amodei, Dario},
 booktitle = {Advances in Neural Information Processing Systems},
 editor = {H. Larochelle and M. Ranzato and R. Hadsell and M.F. Balcan and H. Lin},
 pages = {1877--1901},
 publisher = {Curran Associates, Inc.},
 title = {Language Models are Few-Shot Learners},
 url = {https://proceedings.neurips.cc/paper_files/paper/2020/file/1457c0d6bfcb4967418bfb8ac142f64a-Paper.pdf},
 volume = {33},
 year = {2020}
}

@inproceedings{10.5555/3666122.3667699,
author = {Pourreza, Mohammadreza and Rafiei, Davood},
title = {DIN-SQL: decomposed in-context learning of text-to-SQL with self-correction},
year = {2023},
publisher = {Curran Associates Inc.},
address = {Red Hook, NY, USA},
booktitle = {Proceedings of the 37th International Conference on Neural Information Processing Systems},
articleno = {1577},
numpages = {10},
location = {New Orleans, LA, USA},
series = {NIPS '23}
}

@misc{wang2018robusttexttosqlgenerationexecutionguided,
      title={Robust Text-to-SQL Generation with Execution-Guided Decoding}, 
      author={Chenglong Wang and Kedar Tatwawadi and Marc Brockschmidt and Po-Sen Huang and Yi Mao and Oleksandr Polozov and Rishabh Singh},
      year={2018},
      eprint={1807.03100},
      archivePrefix={arXiv},
      primaryClass={cs.CL},
      url={https://arxiv.org/abs/1807.03100}, 
}

@inproceedings{liu-etal-2022-makes,
    title = "What Makes Good In-Context Examples for {GPT}-3?",
    author = "Liu, Jiachang  and
      Shen, Dinghan  and
      Zhang, Yizhe  and
      Dolan, Bill  and
      Carin, Lawrence  and
      Chen, Weizhu",
    editor = "Agirre, Eneko  and
      Apidianaki, Marianna  and
      Vuli{\'c}, Ivan",
    booktitle = "Proceedings of Deep Learning Inside Out (DeeLIO 2022): The 3rd Workshop on Knowledge Extraction and Integration for Deep Learning Architectures",
    month = may,
    year = "2022",
    address = "Dublin, Ireland and Online",
    publisher = "Association for Computational Linguistics",
    url = "https://aclanthology.org/2022.deelio-1.10/",
    doi = "10.18653/v1/2022.deelio-1.10",
    pages = "100--114"
}

@inproceedings{NEURIPS2020_6b493230,
 author = {Lewis, Patrick and Perez, Ethan and Piktus, Aleksandra and Petroni, Fabio and Karpukhin, Vladimir and Goyal, Naman and K\"{u}ttler, Heinrich and Lewis, Mike and Yih, Wen-tau and Rockt\"{a}schel, Tim and Riedel, Sebastian and Kiela, Douwe},
 booktitle = {Advances in Neural Information Processing Systems},
 editor = {H. Larochelle and M. Ranzato and R. Hadsell and M.F. Balcan and H. Lin},
 pages = {9459--9474},
 publisher = {Curran Associates, Inc.},
 title = {Retrieval-Augmented Generation for Knowledge-Intensive NLP Tasks},
 url = {https://proceedings.neurips.cc/paper_files/paper/2020/file/6b493230205f780e1bc26945df7481e5-Paper.pdf},
 volume = {33},
 year = {2020}
}

@inproceedings{beltagy-etal-2019-scibert,
    title = "{S}ci{BERT}: A Pretrained Language Model for Scientific Text",
    author = "Beltagy, Iz  and
      Lo, Kyle  and
      Cohan, Arman",
    editor = "Inui, Kentaro  and
      Jiang, Jing  and
      Ng, Vincent  and
      Wan, Xiaojun",
    booktitle = "Proceedings of the 2019 Conference on Empirical Methods in Natural Language Processing and the 9th International Joint Conference on Natural Language Processing (EMNLP-IJCNLP)",
    month = nov,
    year = "2019",
    address = "Hong Kong, China",
    publisher = "Association for Computational Linguistics",
    url = "https://aclanthology.org/D19-1371/",
    doi = "10.18653/v1/D19-1371",
    pages = "3615--3620"
}

@article{gemma_2025,
    title={{Gemma 3}},
    url={https://goo.gle/Gemma3Report},
    publisher={Kaggle},
    author={Team Gemma},
    year={2025}
}

@inproceedings{sqlr1,
    title={{SQL}-R1: Training Natural Language to {SQL} Reasoning Model By Reinforcement Learning},
    author={Peixian MA and Xialie Zhuang and Chengjin Xu and Xuhui Jiang and Ran Chen and Jian Guo},
    booktitle={The Thirty-ninth Annual Conference on Neural Information Processing Systems},
    year={2025},
    url={https://openreview.net/forum?id=hgJQcuDwm1}
}

@misc{qwen3,
      title={{Qwen3 Technical Report}}, 
      author={Team Qwen},
      year={2025},
      eprint={2505.09388},
      archivePrefix={arXiv},
      primaryClass={cs.CL},
      url={https://arxiv.org/abs/2505.09388}, 
}

@misc{gpt-5,
      title={OpenAI GPT-5 System Card}, 
      author={Team GPT-5},
      year={2025},
      eprint={2601.03267},
      archivePrefix={arXiv},
      primaryClass={cs.CL},
      url={https://arxiv.org/abs/2601.03267}, 
}

\appendix

\section{Prompts Details}
\label{sec:appendix_prompts}
\mytcbinputwide{prompts/sql2nl.tex}{SQL-2-NL}{0}{bw:domain}{prompt:SQL2NL}
\mytcbinputwide{prompts/FS.tex}{Few-Shot NL2SQL}{0}{bw:domain}{prompt:FS}
\mytcbinputwide{prompts/ZS.tex}{Zero-Shot SQL2NL}{0}{bw:domain}{prompt:ZS}
\mytcbinputwide{prompts/CoT.tex}{Chain of Thoughts NL2SQL}{0}{bw:domain}{prompt:CoT}
\mytcbinputwide{prompts/decomposition.tex}{SQL Query Decomposition}{0}{bw:domain}{prompt:decompose}
\mytcbinputwide{prompts/synth.tex}{Final SQL Synthesizer}{0}{bw:domain}{prompt:synth}
\end{document}